\UseRawInputEncoding
\documentclass[a4paper,10pt,twoside]{article}

\usepackage{clin}       

\usepackage{biblatex}
\usepackage{url}
\usepackage{multicol}
\usepackage{listings}    
\usepackage{xcolor}      
\usepackage{subcaption}
\usepackage{graphicx}
\usepackage{float}
\usepackage{array}
\definecolor{background}{rgb}{0.95,0.95,0.95} 
\definecolor{dkgreen}{rgb}{0,0.6,0}          
\definecolor{mauve}{rgb}{0.58,0,0.82}        

\begin{document}
\lstset{frame=tb,
  language=Python,
  aboveskip=3mm,
  belowskip=3mm,
  showstringspaces=false,
  columns=flexible,
  basicstyle={\small\ttfamily},
  numbers=none,
  numberstyle=\tiny\color{gray},
  keywordstyle=\color{blue},
  commentstyle=\color{dkgreen},
  stringstyle=\color{mauve},
  breaklines=true,
  breakatwhitespace=true,
  tabsize=3,
  backgroundcolor=\color{background}
}

\lstdefinelanguage{json}{
    string=[s]{"}{"},
    stringstyle=\color{blue},
    comment=[l]{:},
    commentstyle=\color{black},
}
\title{Enriching Historical Records:
An OCR and AI-Driven Approach for Database Integration}
\author{
    {\normalsize \bf Zahra Abedi}$^1$ \email{g.z.abedi@gmail.com} \\
    {\normalsize \bf Richard M.K. van Dijk}$^1$ \email{m.k.van.dijk@liacs.leidenuniv.nl} \\
    {\normalsize \bf Gijs Wijnholds}$^1$ \email{g.j.wijnholds@liacs.leidenuniv.nl} \\
    {\normalsize \bf Tessa Verhoef}$^1$ \email{t.verhoef@liacs.leidenuniv.nl}
    \AND \addr{$^1$ Leiden Institute of Advanced Computer Science, Leiden University, Gorlaeus Building, Einsteinweg 55, 2333 CC Leiden, the Netherlands} 
}

\maketitle\thispagestyle{empty} 


\begin{abstract}
This research digitizes and analyzes the Leidse hoogleraren en lectoren 1575-1815 books written between 1983 and 1985, which contain biographic data about professors and curators of Leiden University. It addresses the central question: 'How can we design an automated pipeline that integrates OCR, LLM-based interpretation, and database linking to harmonize data from historical document images with existing high-quality database records?'
We applied OCR techniques, generative AI decoding constraints that structure data extraction, and database linkage methods to process typewritten historical records into a digital format. OCR achieved a Character Error Rate (CER) of 1.08\% and a Word Error Rate (WER) of 5.06\%, while JSON extraction from OCR text achieved an average accuracy of 63\% and, based on annotated OCR, 65\%. This indicates that generative AI somewhat corrects low OCR performance. Our record linkage algorithm linked annotated JSON files with 94\% accuracy and OCR-derived JSON files with 81\%.
This study contributes to digital humanities research by offering an automated pipeline for interpreting digitized historical documents, addressing challenges like layout variability and terminology differences, and exploring the applicability and strength of an advanced generative AI model.
\end{abstract}

\section{Introduction} \label{introduction}

\subsection{Background}\label{background}

Historical documents serve as invaluable sources of information for researchers and historians. Significant research has been dedicated to converting these paper-based records into electronic formats, which helps preserve them, improve accessibility for a broader audience, and discover new links between the data \cite{ooghe2009analysing}.

The process of digitizing documents involves several steps. Initially, the paper-based document is scanned to create a digital image. Subsequently, Optical Character Recognition (OCR) methods are used to extract texts from the scanned documents. Once the OCR-generated texts are acquired, we can apply customized information extraction techniques to interpret the data for our specific needs.

This work is part of the Linking University, City, and Diversity (LUCD) project\footnote{\url{https://www.universiteitleiden.nl/onderzoek/onderzoeksprojecten/wiskunde-en-natuurwetenschappen/liacs-linking-university-city-and-diversity}}, which explores the interactions between the city of Leiden and Leiden University, starting from its foundation in 1575 through the use of data science techniques. A central focus of the project is to determine the impact of the students and scholars on the city. The LUCD project combines data science and history expertise to develop a cohesive platform comprising software architecture with a centralized database and an interactive website front-end. These tools enable users to explore themes such as the origins of Leiden University’s students and professors and the evolution of its academic community over time.\footnote{\url{https://univercity.liacs.nl/dashboard/}}

The proposed approach focuses on enriching a centralized database with high-quality data on students and scholars of Leiden University over the years by integrating the 'Leidse hoogleraren en lecture 1575-1815' books. These books, originally written with a mechanical typewriter between 1983 and 1985 and now available as scanned images\footnote{\url{https://digitalcollections.universiteitleiden.nl/view/item/2078065/pages}}, contain biographical information about scholars and curators from 1575 to 1815. The objectives include designing an automated pipeline to process scanned images, extract text, convert it into a structured format, and harmonize it with existing datasets.

\subsection{Research Questions}\label{research_question}
The main research question is: 'How can we design an automated pipeline that integrates OCR, LLM-based interpretation, and database linking to harmonize data from historical document images with existing high-quality database records?' Sub-questions include:
\begin{itemize}
    \item How can we optimize Tesseract OCR to extract high-accuracy text from scanned historical documents, particularly in cases of inconsistencies and document degradation?
    \item How can we effectively use GPT-3.5 to convert OCR-extracted text into structured JSON while addressing inconsistencies in historical records?
    \item How can we harmonize the structured JSON data into a centralized database with related data?
\end{itemize}

\section{Related Work}\label{related_work}
\subsection{Optical Character Recognition}\label{ocr_related_work}
We used a free and open-source OCR engine, Tesseract, initially developed by Hewlett-Packard in the 1980s and now maintained by Google\footnote{\url{https://en.wikipedia.org/wiki/Tesseract_(software)}}. We trained Tesseract with additional data to enhance its performance for processing historical records and composed a wordlist of the most frequent words to improve the text recognition accuracy, in line with related research by \cite{white2012training}. This study focused on training Tesseract to support Ancient Greek. The paper contains general procedures for customizing Tesseract, including training it with language-specific hints to address challenges from Tesseract's English-oriented design.

\subsection{Information Extraction}\label{information_extraction_related_work}
Information Extraction (IE) is an important early stage in the pipeline for various high-level tasks such as question-answering systems \cite{molla2006named}. Recent research has explored the applicability of large language models (LLMs) like GPT-3.5 for extracting structured information from unstructured text. Research done by \cite{wu2024learning} focuses on how LLMs can be guided to generate structured outputs through schemas and examples without the need for fine-tuning. This approach is especially relevant when dealing with historical records, where data can often be noisy or inconsistent because of OCR and spelling errors in the original paper document.

\subsection{Record Linkage}\label{record_linkage_related_work}
Record linkage, also known as computerized matching, refers to the process of identifying and matching records that correspond to the same entities (such as individuals or businesses) using quasi-identifiers like names, addresses, and dates of birth \cite{winkler2014matching}. \cite{schraagen2014aspects} research the complexities of record linkage within historical databases. His work incorporates methods from various scientific disciplines to tackle different aspects of record linkage, such as handling name variations and exploiting information from related records within the dataset. \cite{mandemakers2023links} presents the LINKS system specifically designed for historical family reconstruction in the Netherlands. This system links historical records to reconstruct the lifelines of persons and family trees, addressing challenges such as the correction of missing or inconsistent data. This work is closely related to our approach because it provides hints for harmonizing historical records.

\section{Data}\label{data}
The 'Leidse hoogleraren en lectoren 1575-1815 DOUSA 80 0211-17' books are written by historian A.A. Bantjes and L. van Poelgeest in 1983 to 1985, published by the Rijksuniversiteit Leiden, Werkgroep Elites 1983-1985. The collection is divided into seven volumes. Volumes 1 to 5 contain information about professors who worked in various faculties of Leiden University, while volumes 6 and 7 contain information on other administrators and staff. While each volume provides information about approximately 57 individuals, the structure remains consistent, including an introduction with a general overview of the professors and their respective faculties, an alphabetical list of names of the individuals mentioned in the volume \footnote{\url{https://digitalcollections.universiteitleiden.nl/view/item/2078129}}, detailed information about each person in alphabetical order \footnote{\url{https://digitalcollections.universiteitleiden.nl/view/item/2078138}}, and a list of historical sources used in the volume, organized numerically \footnote{\url{https://digitalcollections.universiteitleiden.nl/view/item/2078107}}. The detailed information section about each person includes date and place of birth and death, education, career history, additional positions, genealogical details such as family members, and special details like salaries and memberships.

The dataset presents several inconsistencies across its volumes, which we have categorized into distinct types. One prominent issue is incomplete information, where data entries lack uniformity. While some records detail a person's birthplace, birth date, and even death details, others are missing (parts of) this data. For example, figure \ref{fig:inc_1} shows the birth year and birthplace of a person indicated after 'Geb.'('Geb.' refers to 'Geboren'23,, which means 'born' in Dutch.). However, information regarding their death, which is normally mentioned after 'Gest.' ('Gest.,' short for 'Gestorven', meaning 'deceased' in Dutch), is absent.

Another challenge is the variability in formatting within the dataset, where the format of data differs significantly from one entry to another. For example, some individuals' birth dates are recorded precisely as '09-08-1673', while others vary. Figure \ref{fig:inc_2} illustrates this difference, showing a birth year of either 1551 (as noted in sources 2, 6, and 7) or 1554 (as mentioned in source 14). In this example, the death date is either the 16th or 28th (according to source 7) or the 20th (according to source 14), with the death month and year being October 1559.

Furthermore, printing quality across the dataset can vary between volumes. Early editions were printed with lower quality, resulting in blurred or distorted text and images. In contrast, later volumes demonstrate significantly improved printing standards, ensuring clearer and more readable content. The differences in printing quality can impact accessibility, as presented in Figure \ref{fig:inc_3}.\\

\begin{table}[htp]
\centering
\small
\caption{Illustration of Inconsistencies in the Dataset}
\label{inc_info}
\begin{tabular}{|m{2.4cm}|p{12cm}|} 
\hline
\textbf{Inconsistency Type} & \textbf{Example Images} \\ \hline

Incomplete \newline Information & 
\begin{minipage}[t]{0.9\linewidth} 
    \begin{subfigure}[b]{1\textwidth}
        \centering
        \includegraphics[width=\textwidth]{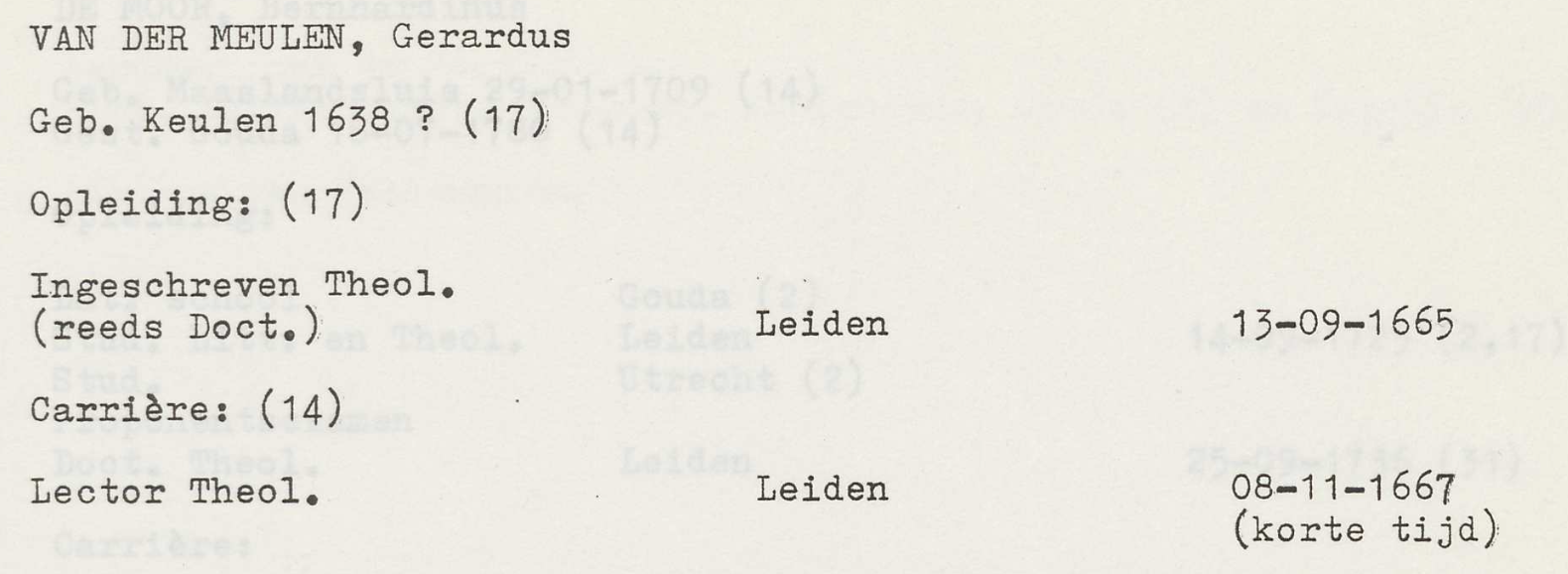}
        \caption{Leidse hoogleraren en lectoren 1575-1815, Volume 1. This record demonstrates the inconsistencies in information completeness, providing details such as birthplace and birth date (mentioned after 'Geb.') but lacking information about the person's death (mentioned after 'Gest.').}
        \label{fig:inc_1}
    \end{subfigure}
\end{minipage} \\ \hline

Variable \newline Formatting & 
\begin{minipage}[t]{0.9\linewidth}
    \begin{subfigure}[b]{1\textwidth}
        \centering
        \includegraphics[width=\textwidth]{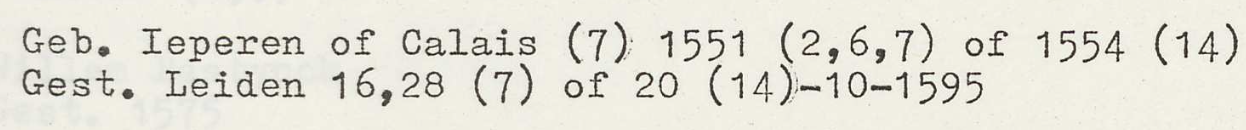}
        \caption{Leidse hoogleraren en lectoren 1575-1815, Volume 1. This record demonstrates the inconsistencies in data formatting, showing birth years as either 1551 (sources 2, 6, and 7) or 1554 (source 14) and death dates as the 16th, 20th, or 28th (sources 7 and 14), with a death month and year of October 1559.}
        \label{fig:inc_2}
    \end{subfigure}
\end{minipage} \\ \hline

Printing \newline Quality & 
\begin{minipage}[t]{0.9\linewidth}
    \begin{subfigure}[b]{1\textwidth}
        \centering
        \includegraphics[width=\textwidth]{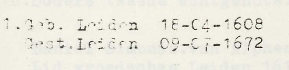}
        \caption{Leidse hoogleraren en lectoren 1575-1815, Volume 1. This record illustrates the low printing quality that might have appeared in the earlier editions.}
        \label{fig:inc_3}
    \end{subfigure}
\end{minipage} \\ \hline

\end{tabular}
\end{table}

\section{Methods}\label{methods}
The research methodology comprises three phases, each addressing a sub-question to examine the problem systematically and thoroughly examine the problem and evaluate it in detail. The phases are:
\begin{itemize}
    \item \textbf{Phase 1:} PDF to PNG Conversion, Image Preprocessing, and Optical Character Recognition (OCR) \& Text Segmentation Per Person
    \item \textbf{Phase 2:} Text to JSON Extraction with AI
    \item \textbf{Phase 3:} Record Linkage and Database Enrichment
\end{itemize}
 The code used for these methods can be found in the GitHub repository.\footnote{GitHub repository: \url{https://github.com/Leiden-University-City-Lab/BantjesAdapter.git}}

\subsection{Phase 1: PDF to PNG Conversion, Image Preprocessing, and OCR \& Text Segmentation Per Person}

This phase addresses the first sub-question: \textit{How can we extract high-accuracy text from scanned historical documents using OCR techniques?} The focus is on preparing individual text files from the original PDFs. The steps involved are:
\begin{enumerate}

    \item \textbf{PDF to PNG Conversion}: The first step is to prepare the right format for processing. We downloaded the PDF scans from the website\footnote{\url{https://digitalcollections.universiteitleiden.nl/view/item/2078065/pages}}. Since Tesseract OCR doesn't support PDFs, we converted them to an image format. Among the compatible formats, we chose PNG because it offers efficient compression without losing image quality, making it more suitable than TIFF, which tends to result in larger file sizes. To convert the PDF to PNG, we use the Python module pdf2image\footnote{\url{https://pypi.org/project/pdf2image/}}.
    
\item \textbf{Image Preprocessing}: Preprocessing the scanned images is essential to improve OCR accuracy, particularly for historical documents that may be degraded. While Tesseract OCR uses the Leptonica library for internal image processing, its built-in operations can sometimes fall short, reducing accuracy\footnote{\url{https://tesseract-ocr.github.io/tessdoc/ImproveQuality.html}}. To address this, additional preprocessing is applied to enhance the quality of these historical records. The preprocessing steps we plan to use are:
\begin{enumerate}
    \item \textbf{Denoising:} Reducing noise, such as specks, smudges, or distortions, is essential to improving OCR accuracy. Noise can interfere with character recognition, affecting the entire text-processing pipeline. For this, an OpenCV-based method was applied, using parameters optimized to minimize noise while preserving image details.
\begin{lstlisting}[language=Python, caption={Denoising the images using OpenCV}]
# Denoise image with OpenCV
denoise = cv2.fastNlMeansDenoisingColored(image, None, 5, 5, 7, 21)
\end{lstlisting}

    \item \textbf{Grayscale:} Grayscaling transforms a continuous tone image into varying shades of gray. By converting a colored image to grayscale, it can often reveal more details than the original colored version. During preprocessing, we converted the images to grayscale using OpenCV.
\begin{lstlisting}[language=Python, caption={Converting the images to grayscale using OpenCV}]
# Convert image to grayscale
gray = cv2.cvtColor(denoise, cv2.COLOR_BGR2GRAY)
\end{lstlisting}
    \item \textbf{Binarization:} Binarization simplifies a grayscale image by converting it into a binary format, setting pixels above a specific threshold to white and those below to black. This step enhances OCR accuracy by reducing noise and clarifying character boundaries. For our process, a threshold value of 200 was used to ensure optimal separation of text from the background.
    \begin{lstlisting}[language=Python, caption={Binarizing the images using OpenCV}]
# Grayscale to binary with OpenCV threshold
th, image = cv2.threshold(gray, 200, 255, cv2.THRESH_BINARY)
\end{lstlisting}
\end{enumerate}
    \item \textbf{Optical Character Recognition (OCR) \& Text Segmentation Per Person}: This step extracts text from images and segments it by individual persons, creating separate text files for clarity and subsequent processing.\\\\
    \textbf{Optical Character Recognition (OCR):} After preprocessing the images, Tesseract OCR was used for text extraction. Tesseract, an open-source tool, allows parameter customization for improved performance with different fonts, styles, and layouts. Here are the essential aspects considered:
\begin{itemize}
    \item Training: We trained Tesseract using a dataset created from screenshots of different lines of text from our dataset. For each screenshot, we generated a corresponding .txt file with the correct text. After training, a trained language data file, \texttt{.traineddata}, was created to support the recognition of specific language elements and improve overall accuracy. However, the documentation notes that retraining Tesseract may not improve results unless a very unusual font is used\footnote{\url{https://tesseract-ocr.github.io/tessdoc/ImproveQuality.html}}.
    \item Language Configuration: Tesseract can recognize text in multiple languages. Dutch was specified as the recognition language to adjust the character set and dictionary for precise results since the texts were produced between 1981 and 1983.
    \item Page Segmentation Modes (PSM): Tesseract provides 14 different PSMs, each optimized for various layout scenarios. After experimenting with different PSM settings, we realized that PSM 4, which processes text in single columns, works best for our documents.
\end{itemize}
\textbf{Text Segmentation Per Person:} To manage errors and improve file organization, the text is split into individual .txt files for each person. An algorithm using regular expressions in Python was developed based on specific patterns in the book. Each person's last name, which is the only capitalized word on the first line of the page, marks the start of a new file. The algorithm sorts PNG files numerically, iterates through them, performs OCR to extract text, and checks for last names. When a last name is found, a new .txt file is created for that person. If no last name is detected, the information is added to the most recent .txt file. This process continues until all pages are processed. For an example of a generated file, see Appendix \ref{appendixF}.
 \end{enumerate}
\subsection{Phase 2: JSON Extraction from Text with AI}
After splitting the text into separate files for each person, the next step is to extract relevant data into a structured format. JSON was chosen for its simplicity and readability, both for humans and machines. Initially, we explored using regular expressions in Python to extract information following specific titles such as 'Opleiding' (Education) and 'Loopbaan' (Career) from the text. However, OCR errors led to unreliable results, so we shifted to using an AI model. 

\subsubsection{Schema Definition}
To generate a consistent output that contains all the necessary fields, we need to define a schema for our model. Given the complexity of our desired output, which involves highly nested and extensive JSON files, this task is challenging. To address this, we use Pydantic\footnote{\url{https://pydantic.dev}}, a Python library that allows us to define the keys of our JSON as classes. Pydantic simplifies data validation with Python type annotations and ensures the output matches the defined schema without complex error-checking code. It also automatically generates JSON schemas\footnote{\url{https://docs.pydantic.dev/latest/concepts/json_schema/}}, making the structure more manageable. The full JSON schema can be found in our GitHub repository\footnote{\url{https://github.com/Leiden-University-City-Lab/BantjesAdapter/blob/main/AI/json_schema.json}}. Below is an example of a Pydantic class for extracting career-related information:\\ 
\begin{lstlisting}[language=Python, caption={Example Pydantic Class for Career Information}]
class Career(BaseModel):
    %Identifying information about the person’s career.
    job: Optional[str] = Field(None, description='The type of job', examples=['Hoogleraar Geschiedenis'])
    location: Optional[str] = Field(None, description='The location of the job', examples=['Leiden'])
    date: Optional[str] = Field(None, description='The date of the job.', examples=['1601-10-20', '1601'])
    source: Optional[str] = Field(None, description='The source of the info mentioned in parentheses', examples=['6'])

class Person(BaseModel):
    FirstName: str = Field(..., description="The first name of a person", examples=['Cornelis', 'Johannes'])
    LastName: str = Field(..., description="The last name of a person", examples=['EKAMA'])
    BirthDate: Optional[str] = Field(None, description="Birth date, Usually found after Geb.", examples=['1601-10-20', '1601', '1601-10'])
    careers: List[Career]
\end{lstlisting}

\subsubsection{Extraction Techniques}
To extract the information, we use the GPT-3.5 Turbo model from OpenAI platform\footnote{\url{https://platform.openai.com/docs/models/gpt-3-5-turbo}} as our Large Language Model (LLM). GPT-3.5 Turbo's function-calling feature enables us to define functions in an API call and have the model intelligently output a JSON object with the necessary key-value pairs. Instead of executing the function, the Chat Completions API produces JSON that can be used to call the function in our code. This approach ensures that structured data is returned more reliably\footnote{\url{https://platform.openai.com/docs/guides/function-calling}}. OpenAI's function calling capabilities, combined with Pydantic, allow for a more structured and reliable approach to output parsing. Pydantic's data models are defined with familiar Python-type annotations, making them intuitive and easy to use. The model’s output is always expected to conform to the Pydantic schema. If the required fields like the ‘last name’ are missing, Pydantic will raise an error. If such an error were to occur, the prompt would be re-executed until the model returned a valid JSON structure according to the schema.
That said, while the schema ensures the correct structure, the degree to which values are filled for each JSON key may differ across prompts. Some fields may have richer data depending on how much information the model could extract from the text.

The prompt we used is shown in appendix \ref{appendixB}, where the \texttt{model} attribute specifies the GPT model to use. The \texttt{messages} list defines the context, with the \texttt{system} message giving instructions and the \texttt{user} message containing the input data. The \texttt{response\_model} attribute ensures that the output matches the Pydantic schema, and the \texttt{tool\_choice} attribute directs the model's decision-making. The \texttt{person\_info} is OCR-obtained input text, passed through the \texttt{user} message content. The process is designed to maximize output reliability with minimal retries. We used the default temperature setting of 0.7, which introduces more randomness in the model’s output. In our context, this flexibility is useful as it allows the model to correct potential OCR-related errors. Another important hyperparameter is \texttt{max\_retries=3}. This setting defines how many times the system will retry in case of a failed API call. Appendix \ref{appendixG} shows an example of a generated JSON file.

\subsection{Phase 3: Record Linkage \& Database Enrichment}
After extracting the data into JSON files, the next step is to enrich the centralized database with this information. The centralized database, developed by the LUCD project, already contains high-quality data on professors and students associated with Leiden University. The original ER diagram, designed by \cite{koning2022}, provides the foundation for this database structure. However, the original structure was insufficient to encapsulate all our data.

To address this, we updated the schema by adding tables for 'education,' 'career,' and 'particularity,' as well as three new columns to the 'person' table: 'AlternativeLastName,' 'rating,' and 'faculty.' The 'rating' column helps identify data quality: 3 for high-quality original data, 2 for matching existing entries, and 1 for new entities. This system ensures we can filter and check new data for errors. High-rated records will not be overwritten; instead, we will enrich them if their corresponding tables are empty.

\subsubsection{Linking Algorithm}
Quasi-identifiers such as names, addresses, and birth dates are essential for record linkage, as they help uniquely identify individuals \cite{winkler2014matching}. To compare the values between two fields, we allow for partial matches by breaking names into substrings and checking if parts of the first or last name appear in the database records. This approach identifies matches like 'Casper Janszoon' and 'Casper Johannes'.
When linking two records together, we assess them against two specific conditions. If either condition is satisfied, we will consider the records to match and integrate any additional information from our JSON files into these records. The conditions are as follows:

\textbf{First condition:} the first name and last name must match, and either the birth year must be the same or the birth city must match.

\textbf{Second condition:} the last name and birth year must match, and either the birth city or the birth country must match.

These conditions are based on commonly available identifiers. Adding more criteria could reduce matches, as some records may lack complete information.

We also introduce a condition for uncertain matches, where the data does not perfectly align but there is still a potential match. For these cases, the algorithm identifies records where the birth year and birthplace do not match, but the names match. If such a match is found, we create a new person in the database but introduce a relation between this newly added person and the person from the database who we thought might be the same individual. This approach allows us to easily identify uncertain cases and review them further.

\section{Evaluation}\label{results}
We evaluated the performance of each phase in our methodology, measuring the performance per phase separately and overall. For this evaluation, we composed a sample set of 10\% of the total number of persons in the dataset. The sample set was chosen to represent a diverse range of data across all volumes.

\subsection{Phase 1 Evaluation: Quality Assessment of Generated Text }\label{phase1_evaluation}
In our evaluation of Phase 1, we assessed the quality of the generated text. To ensure a comprehensive evaluation, we created a labeled dataset comprising 10\% of the total text files from all volumes. This labeled dataset served as our ground truth for comparison with the OCR-generated text.

In evaluating the quality of the text recognition system, Character Error Rate (CER) and Word Error Rate (WER) have been chosen as the primary metrics. These metrics are widely accepted in the scientific community for assessing OCR systems, as they provide a clear and quantifiable measure of errors at both the character and word levels \cite{leifert2019end}.

In contrast, metrics such as precision, recall, and F-measure, commonly used in information retrieval through the bag-of-words (BOW) model, were considered less suitable because this model suffers from several drawbacks with the main drawback being that the model does not consider the reading order of words, failing to penalize permutations of recognized words \cite{leifert2019end}.

Given this context, CER and WER have been used to evaluate the OCR system.

To calculate these metrics, an algorithm based on the Levenshtein distance has been implemented. The algorithm processes pairs of files: one from the OCR-generated text files and one from the correct text files that we manually labeled. It computes the edit distance between the two texts and normalizes it by the length of the reference text, thus obtaining CER and WER.

The steps involved in the algorithm are as follows:
\begin{itemize}
    \item Normalization: both texts are stripped of leading and trailing whitespaces and converted to lowercase to ensure consistency in comparison.
    \item Levenshtein Distance Calculation: the edit distance is calculated using a dynamic programming approach that fills a matrix with the minimum number of edits (insertions, deletions, substitutions) required to transform one string into another.
    \item Metric Computation: both CER and WER are derived from the Levenshtein distance. CER and WER normalize this distance by dividing the total number of edits by the number of characters and words in the reference text, respectively, providing a standardized error rate.
\end{itemize}

\begin{figure}[htp]
    \centering
    \begin{subfigure}[b]{0.45\textwidth}
        \centering
        \includegraphics[width=\linewidth]{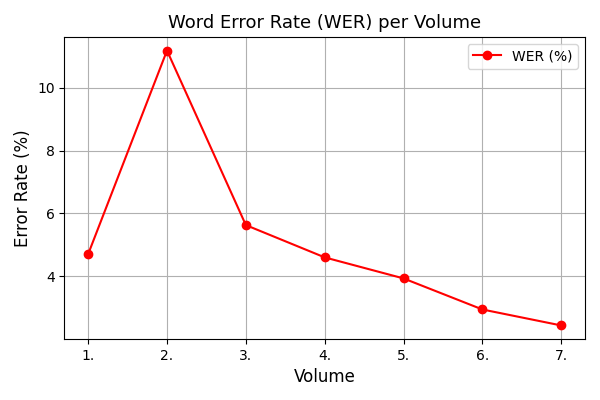}
        \caption{Average Word Error Rate (WER) per Volume.}
        \label{fig:wer}
    \end{subfigure}
    \hfill
    \begin{subfigure}[b]{0.45\textwidth}
        \centering
        \includegraphics[width=\linewidth]{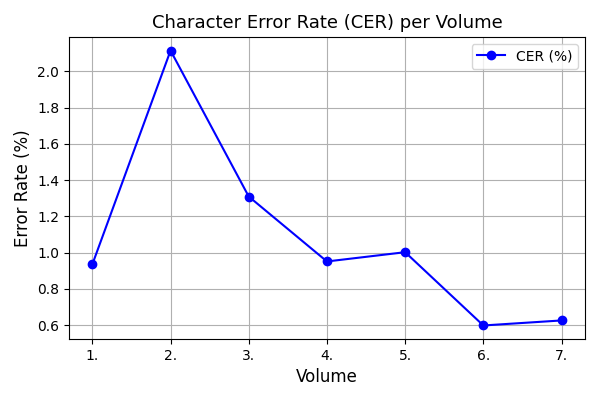}
        \caption{Average Character Error Rate (CER) per Volume.}
        \label{fig:cer}
    \end{subfigure}
    \caption{Comparison of Average Word Error Rate (WER) and Character Error Rate (CER) Across Different Volumes}
    \label{fig:error_rates}
\end{figure}

CER\footnote{\url{https://en.wikipedia.org/wiki/Word_error_rate}}, defined as the ratio of the sum of insertions, deletions, and substitutions to the total number of characters in the reference text, offers an inverted measure of character accuracy. WER\footnote{\url{https://docs.kolena.com/metrics/wer-cer-mer}} extends this concept to the word level, providing insights into how well entire words are recognized. 
The evaluation involved comparing the OCR-generated text with our labeled dataset to compute the Word Error Rate (WER) and Character Error Rate (CER). We then averaged these rates for each volume. The results are presented in Figure \ref{fig:error_rates}.\\

While WER is greater than CER across all volumes, their relative difference differs since it is influenced by how character errors are distributed. In the WER diagram for Volume 2, we observe a downward trend, whereas the CER diagram exhibits a fluctuating pattern. This variation indicates that some volumes expose a more homogeneous error distribution, where errors are evenly spread across words, while others show a more heterogeneous distribution, with errors concentrated in specific words.

Among these volumes, volume 2 exhibits significantly higher error rates than the others, with WER and CER both notably elevated. This suggests that the OCR system encountered more difficulties with the text in this particular volume. The primary factor contributing to the poorer performance observed in volume 2 is likely the quality of printing. Poor print quality can introduce noise and distortions in the text, making it harder for the OCR system to accurately recognize characters and words. For instance, faded ink, smudges, or uneven text alignment can significantly impact recognition accuracy. 

Our analysis of the OCR output revealed that errors were not significantly influenced by using Dutch documents with an English-centric OCR tool. Instead, inaccuracies were primarily caused by document quality issues, such as faded ink, smudges, and low-resolution scans, which led to unclear text extraction. Additionally, variations in word positioning, especially near page borders or within complex layouts, further contributed to recognition errors.

Common OCR errors included the substitution of characters, such as the lowercase ‘l’ being misread as ‘1’ or ‘i’, and the omission of characters, particularly when letters were faded or closely spaced, like ‘Leiden’ being read as ‘Lden’.

\subsection{Phase 2 Evaluation:  Quality Assessment of Generated JSON}\label{phase2_evaluation}
In Phase 2 of our evaluation, we aimed to assess the quality of the JSON files created by our AI model. For 10\% of the individuals in our dataset, we manually created correct JSON files that align with our predefined schema. These files were checked against the original images to ensure their accuracy.\\\\
To assess the impact of OCR-related errors and text quality on the generated JSON files, we created two sets of JSON files for accuracy evaluation using GPT-3.5 Turbo as our AI model. The first set used manually corrected text inputs. For each text file, we generated five JSON files to calculate average accuracy, ensuring a comprehensive assessment and avoiding non-representative outcomes. The second set was derived from OCR-generated text inputs. For each text file in this set, we also generated five JSON files. Here, the objective was to gauge accuracy considering potential spelling errors and OCR inaccuracies.\\\\
To evaluate these JSON files, we conducted a comparison between the AI-generated JSON files and the correct JSON files:
\begin{enumerate}
    \item Normalization by lowercasing: to ensure a consistent comparison, we converted all text values to lowercase. This step helped in making the comparison case-insensitive, thereby focusing only on content accuracy rather than case differences.
    \item Value comparison: we compared each key-value pair in the AI-generated JSON files with the corresponding pair in the correct JSON files. This comparison was performed separately for two sets of AI-generated JSON files: those based on Correct Text Files and those based on OCR-Generated Text Files.
    \item Accuracy assessment: for each key, we calculated the accuracy by determining the percentage of correct values generated by the AI.
    \item Key categorization: keys in the JSON files were categorized into meaningful groups such as 'Main person', 'Education', 'Careers', 'Particularities', 'Spouses', 'Parents', 'Grandparents', 'In-laws', 'Children', and 'Far family'. This categorization allows us to perform a focused analysis of accuracy and helps us detect the error-prone areas in our JSON files.
\end{enumerate}
Below is a comparison example of a small section from JSON files, focusing on keys related to the 'Main person' category. The first JSON is manually created and therefore is correct, while the second JSON is generated using our AI model. In our dataset, the \texttt{type\_of\_person} attribute, set to 1, denotes that the individual is classified as a professor. Other types of persons that appear in the database include students and curators.\\ 
\begin{lstlisting}[language=json,firstnumber=1,  caption={Example of a correct JSON file containing the key-value pairs related to the 'Main person' category}]
{
  "first_name": "Caspar Janszoon",
  "last_name": "COOLHAES",
  "affix": null,
  "gender": "Man",
  "alternative_last_names": ["KOOLHAES", "KOOLHAAS", "COELAES"],
  "type_of_person": 1, 
  "faculty": "Theologie",
  "birth_country": "Duitsland",
  "birth_city": "Keulen",
  "birth_date": "1534-01-24",
  "death_date": "1615-01-15",
  "death_city": "Leiden"
}
\end{lstlisting}

\begin{lstlisting}[language=json,firstnumber=1, caption={Example of a JSON file, made using our AI model, containing the key-value pairs related to the 'Main person' category}]
{
  "first_name": "Caspar Janszoon",
  "last_name": "COOLHAES",
  "affix": null,
  "gender": "Man",
  "alternative_last_names": [],
  "type_of_person": 1,
  "faculty": "Theologie",
  "birth_country": null,
  "birth_city": "Keulen",
  "birth_date": "1534",
  "death_date": "1615",
  "death_city": "Leiden"
}
\end{lstlisting}
Table \ref{tab:accuracy} shows the comparison results of the two JSON files above. The 'Accuracy' score represents the percentage of correct values generated by the AI for each key in the JSON file. For example, the \texttt{first\_name} key has an accuracy of 100.00\%, indicating that the AI correctly generated the first name in all instances. 
In our evaluation, if the AI extracts '1615' but the correct complete date is '1615-01-15', we consider it inaccurate even if the year information matches. This approach ensures precision in our accuracy assessment by mandating exact date formats where necessary.
\begin{table}[htp]
\small
\centering
\begin{tabular}{|l|l|c|}
\hline
\textbf{Category} & \textbf{Key} & \textbf{Accuracy}\\ \hline
Main Person       & first\_name    & 100.00\% (1/1)               \\ \hline
Main Person       & last\_name     & 100.00\% (1/1)\\ \hline
Main Person       & affix        & 100.00\% (1/1)\\ \hline
Main Person       & gender       & 100.00\% (1/1)\\ \hline
Main Person       & alternative\_last\_names & 0.00\% (0/3)\\ \hline
Main Person       & type\_of\_person & 100.00\% (1/1)\\ \hline
Main Person       & faculty      & 100.00\% (1/1)\\ \hline
Main Person       & birth\_country & 0.00\% (0/1)\\ \hline
Main Person       & birth\_city    & 100.00\% (1/1)\\ \hline
Main Person       & birth\_date    & 0.00\% (0/1) \\ \hline
Main Person       & death\_date    & 0.00\% (0/1)\\ \hline
Main Person       & death\_city    & 100.00\% (1/1)\\ \hline
\end{tabular}
\caption{Example of accuracy scores for the main person category in the JSON file.}
\label{tab:accuracy}
\end{table}

\begin{figure}[htp]
    \centering
    \includegraphics[width=\textwidth]{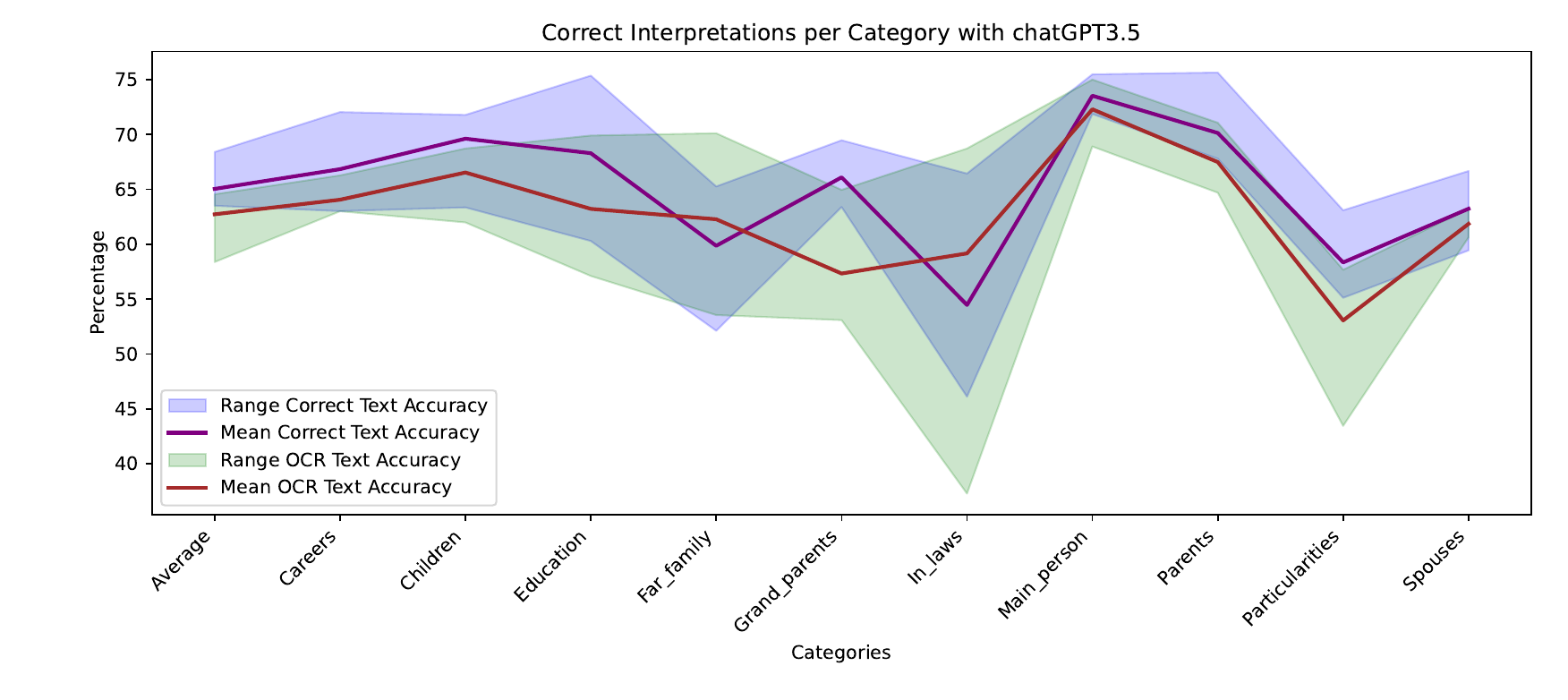}
    \caption{Correct interpretations per category using ChatGPT-3.5 in five independent trails. The chart displays the accuracy min-max ranges and means for JSON files generated from both correct text files and OCR-generated text files across various categories.}
    \label{fig:correct_interpretations_chatgpt35}
\end{figure}
Figure \ref{fig:correct_interpretations_chatgpt35} illustrates the accuracy of JSON file generation across various categories, comparing results obtained from correct text files and OCR-generated text files. It highlights the mean accuracy and min-max range of correct interpretations of five trails for each category, demonstrating the performance differences between the two text sources.

The average accuracy scores were higher for JSON files generated from correct text inputs compared to those generated from OCR-generated inputs. This discrepancy highlights the impact of OCR errors, including spelling mistakes and incorrectly recognized words, on the AI’s performance. The 'Main person' category demonstrated the highest accuracy scores, likely due to the uniformity and prominent placement of this information at the top of each page.

Conversely, categories such as 'In-laws' and 'Far family' exhibited lower accuracy scores, particularly for JSON files created from correct text inputs. This may be attributed to the inconsistent formatting of these sections in the dataset. For instance, information about in-laws was sometimes included within the spouse's section or presented in a dedicated section. Such variability posed challenges for the AI, leading to inaccuracies.

Interestingly, JSON files generated from OCR text occasionally outperformed those based on correct text in specific categories like 'In-laws'. This anomaly could be due to the AI leaving certain fields blank when working with error-prone OCR text, while attempts to extract more relationships from correct text inadvertently resulted in inaccuracies.

The 'Particularities' category also showed significantly lower accuracy scores, primarily due to the complexity of the information and its susceptibility to OCR errors. Overall, the evaluation emphasized the importance of accurate input text for generating high-quality JSON files.

A noteworthy point is that the initial OCR process with Tesseract introduced some errors, particularly with city names, due to document quality and printing inconsistencies. These were partially mitigated through the use of the GPT-3.5 Turbo model for data extraction. The LLM demonstrated an ability to recognize and correct certain inconsistencies, especially for commonly known city names where the OCR had misread characters. For instance, misinterpreted names like ’Roterdam’ instead of ’Rotterdam’ were correctly identified during the extraction phase. We also noticed that in some cases, the model was able to correct OCR errors in dates based on contextual information. For example, an incorrect OCR output of '1897-1598' was adjusted to the correct version '1597-1598'. This suggests that the language model’s contextual understanding contributed to refining and improving the overall data quality. However, this correction was not universal.

\subsection{Phase 3 Evaluation: Quality Assessment of Linking Algorithm}\label{phase3_evaluation}
To evaluate the enrichment algorithm, we used sample data consisting of two sets of JSON files. The first set was generated using OCR text, and the second set was manually created to ensure correctness. By comparing these two sets of JSON files, we can better understand the impact of potential OCR errors and the errors introduced by the AI model on the record-linking process.\\\\
To measure the performance, we executed our algorithm on these two sets of JSON files and calculated the results for each volume. The following is the schema of the JSON file created specifically for our evaluation, which contains information on the performance of our database enrichment algorithm. Each JSON file contains objects corresponding to an individual from our dataset. If the linking algorithm finds a person in the database, the \texttt{new\_person} attribute is set to false, and the \texttt{person\_id} represents the unique ID of the identified person from the database. If no person is found, a new person will be added to the database, and therefore, the \texttt{new\_person} attribute is set to true, and the \texttt{person\_id} shows the ID of the newly added record. The \texttt{maybe\_same\_person} attribute is set to true when there is insufficient evidence for the linking algorithm to match two records definitively but identifies a potential match. In such cases, a new record is created under the ID specified by \texttt{person\_id}.

\begin{lstlisting}[language=json,firstnumber=1, caption={JSON schema for evaluating database enrichment}]
{
  "name_of_file": {
    "person_id": int,
    "new_person": bool,
    "maybe_same_person": bool
  }
}
\end{lstlisting}

The accuracy results of the linking algorithm using correct JSON files are shown in Table \ref{tab:acc_db_correct}, and Table \ref{tab:acc_db_ocr} shows the results using OCR-generated JSON files. The 'Person\_ID Accuracy' measures how often the algorithm correctly identifies existing individuals. Using correct JSON files, accuracy ranges from 71.43\% to 100\%. With OCR-generated files, accuracy drops, ranging from 57.14\% to 100\%, indicating OCR errors impact the algorithm's ability to identify individuals correctly. The 'New\_Person Accuracy' measures how accurately the algorithm identifies new individuals. Using correct JSON files, accuracy is generally high (85.71\% to 100\%). With OCR-generated files, accuracy is lower, ranging from 57.14\% to 100\%. The 'Maybe\_Same\_Person Accuracy' measures how often the algorithm correctly flags uncertain matches. This metric consistently shows high accuracy (75\% to 100\%). The comparison shows the impact of OCR errors on the linking algorithm. While the algorithm performs well with manually created JSON files, its accuracy decreases with OCR-generated files, particularly for identifying existing individuals and new entries.\\

\begin{table}[htp]
\small
\centering
\begin{tabular}{|l|c|c|c|c|}
\hline
\textbf{Volume} & \textbf{Accuracy} & \textbf{Accuracy} & \textbf{Accuracy} & \textbf{Accuracy} \\
& \textbf{Person\_ID} & \textbf{New\_Person} & \textbf{Maybe\_Same\_Person} & \textbf{Average} \\
\hline
Volume 1 & 71.43\% & 85.71\% & 100\% & 85.71\% \\
\hline
Volume 2 & 80\% & 80\% & 100\% & 86.67\% \\
\hline
Volume 3 & 100\% & 100\% & 100\% & 100\% \\
\hline
Volume 4 & 100\% & 100\% & 100\% & 100\% \\
\hline
Volume 5 & 87.5\% & 87.5\% & 100\% & 91.67\% \\
\hline
Volume 6 & 100\% & 100\% & 75\% & 91.67\% \\
\hline
Volume 7 & 100\% & 100\% & 100\% & 100\% \\
\hline
\textbf{Total} & 91.28\% & 93.32\% & 96.43\% & 93.67\% \\
\hline
\end{tabular}
\caption{Accuracy scores of the linking algorithm using correct JSON files}
\label{tab:acc_db_correct}
\end{table}

\begin{table}[htp]
\small
\centering
\begin{tabular}{|l|c|c|c|c|c|}
\hline
\textbf{Volume} & \textbf{Accuracy} & \textbf{Accuracy} & \textbf{Accuracy} & \textbf{Accuracy} \\
& \textbf{Person\_ID} & \textbf{New\_Person} & \textbf{Maybe\_Same\_Person} & \textbf{Average} \\
\hline
Volume 1 & 57.14\% & 57.14\% & 100\% & 71.43\% \\
\hline
Volume 2 & 60\% & 60\% & 80\% & 66.67\% \\
\hline
Volume 3 & 83.33\% & 83.33\% & 100\% & 88.89\%\\
\hline
Volume 4 & 66.67\% & 66.67\% & 100\% & 77.78\%\\
\hline
Volume 5 & 62.5\% & 62.5\% & 100\% & 75\% \\
\hline
Volume 6 & 100\% & 100\% & 75\%  & 91.67\%\\
\hline
Volume 7 & 100\% & 100\% & 85.71\%  & 95.24\%\\
\hline
\textbf{Total} & 75.66\% & 75.66\% & 91.53\% & 80.95\% \\
\hline
\end{tabular}
\caption{Accuracy scores of the linking algorithm using JSON files made from OCR-generated text files}
\label{tab:acc_db_ocr}
\end{table}

Table \ref{tab:acc_db_comp} compares the counts of new persons generated by the linking algorithm across different volumes when using correct JSON files versus OCR-generated JSON files. The table includes the number of instances in each volume, the correct count of new persons that should have been created, and the actual counts generated by the algorithm for both sets of JSON files.\\
In volumes 6 and 7, a significantly higher number of new people were generated compared to volumes 1 to 5. This difference is because the central database only contains information on professors and students of Leiden University while volumes 6 and 7 contain data on curators of Leiden University. As a result, the linking algorithm was unable to find matches for curators because they were not present in the centralized database, thus identifying more new individuals in these volumes. 

\begin{table}[htp]
\small
\centering
\begin{tabular}{|l|c|c|c|c|c|}
\hline
\textbf{Volume} & \textbf{Number of} & \textbf{Correct} & \textbf{Generated} & \textbf{Generated} \\
 & \textbf{Persons} & \textbf{New\_Person Count} & \textbf{New\_Person Count} & \textbf{New\_Person Count} \\
  &  &  & \textbf{(Correct JSON)} & \textbf{(OCR JSON)} \\
\hline
Volume 1 & 7 & 0 & 1 & 3 \\
\hline
Volume 2 & 5 & 0 & 1 & 2 \\
\hline
Volume 3 & 6 & 1 & 1 & 2 \\
\hline
Volume 4 & 3 & 1 & 1 & 2 \\
\hline
Volume 5 & 8 & 2 & 3 & 5 \\
\hline
Volume 6 & 4 & 3 & 3 & 3 \\
\hline
Volume 7 & 7 & 7 & 7 & 7 \\
\hline
\end{tabular}
\caption{Comparison of the number of new persons in the sample data identified by the linking algorithm across correct JSON and OCR-generated JSON files}
\label{tab:acc_db_comp}
\end{table}

\section{Discussion}\label{discussion}
In this study, our focus has been on enhancing the quality of OCR-generated text, optimizing information extraction using AI models, and refining the performance of our record linkage algorithm. However, several persistent challenges continue to impact the accuracy and comprehensiveness of our results, highlighting the need for ongoing refinement.\\\\ Despite efforts to enhance OCR text quality, residual issues impact downstream processes, particularly in JSON generation. Variability in page layouts across our dataset poses a challenge. Instances where information from separate columns is interpreted as a single line disrupt context continuity and coherence, potentially leading to inaccuracies in JSON outputs.\\\\
During information extraction, we employed a uniform JSON schema across all volumes, assuming document structure consistency. However, subtle variations between volumes, such as different terminologies ('Carrière' vs. 'Loopbaan' for job information), suggest that volume-specific JSON schemas might enhance model accuracy. Tailoring schemas to unique volume characteristics could improve the contextual relevance and reliability of extracted data.\\\\
In some cases, certain information appears in the data that we have not considered as separate JSON keys. For instance, a person’s salary is sometimes mentioned in the 'Particularities' section ('Bijzonderheden' in Dutch). The reason for not including these as fixed JSON keys is that they are not consistently mentioned for all individuals, which would result in many null values. However, these specific details can be added to the JSON schema if they frequently appear and are deemed necessary.\\\\
The linking algorithm in this study uses specific criteria such as first name, last name, birth city, and birth year to link records. Our approach employs partial matching for names to accommodate variations within names. While first names such as 'Casper Janszoon' and 'Casper Johannes' are considered a match, it does not accommodate variations such as ‘Coolhaas' vs. 'Koolhaes' due to stricter matching rules. This decision was made to reduce the number of incorrect matches while maintaining precision. Implementing techniques like Levenshtein distance could enhance the algorithm's flexibility, allowing for less strict matching criteria and potentially reducing false negatives.\\\\
For our evaluation, we selected a sample consisting of 10\% of the total people in the dataset, amounting to 40 individuals. We did not choose a larger sample because manually creating the ground truths for the evaluations was an exhaustive and time-consuming task. However, from a statistical point of view, to achieve a confidence level of 95\% with a 5\% error margin, the ideal sample size for a population of 400 individuals would be 196. This represents approximately 49\% of the total population. Calculations were performed using an online sample size calculator, which takes into account the population size, margin of error, and desired confidence level\footnote{\url{https://www.qualtrics.com/uk/experience-management/research/determine-sample-size/?rid=ip&prevsite=en&newsite=uk&geo=NL&geomatch=uk}}.\\\\
Finally, although precision and recall as performance measures can be informative and beneficial for the linking stage, we emphasized accuracy as our primary metric because it provides a general overview of the extraction phase, considering *all* types of errors in a single measure. The benefit of precision and recall measures seems to be weak in our specific situation. Due to the small dataset, a person's chance of linking with the wrong person stored in the database is low. This did not happen much during our experiments. More often, two records about the same person were not matched due to a lack of information, this person was added as a new person to the database with a lower quality rate to indicate the need for manual correction, mitigating these false positives.

\section{Future Work}\label{future_work}
This study introduces an approach to digitizing historical data using OCR and AI technologies. Moving forward, several avenues for future research can improve the sophistication and efficiency of these methodologies. Future research could explore using advanced multi-modal AI models, such as GPT-4o, developed by OpenAI\footnote{\url{https://platform.openai.com/docs/models/gpt-4o}}. GPT-4o can handle text and image inputs to generate structured text outputs, making it suitable for automated data extraction from historical documents. By using GPT-4o's capabilities, there is potential to simplify workflows and improve accuracy in generating JSON-formatted outputs directly from document images, potentially eliminating OCR steps.\\
Additionally, the study identified challenges in the current linking algorithm, particularly in accommodating variations within names. Implementing techniques like Levenshtein distance could enhance the algorithm’s flexibility, allowing for less strict matching criteria and potentially reducing false negatives. \\
Future research should also consider the use of volume-specific JSON schemas to account for subtle differences in document structures across volumes. Furthermore, exploring prompt engineering could lead to better results in information extraction. By refining and experimenting with different prompts, future research could optimize the performance of AI models in generating accurate and comprehensive JSON outputs.

\section{Conclusion}\label{conclusion}
This study tackled the challenge of accurately extracting and transforming historical records into a centralized digital database using a three-phase methodology. Each phase targeted a specific aspect of the digitization pipeline, addressing key sub-questions.\\\\
In the first phase, we focused on extracting high-accuracy text from scanned historical documents using OCR techniques. By employing Tesseract OCR and applying image preprocessing methods such as denoising and binarization, we achieved a Character Error Rate (CER) of 1.08\% and a Word Error Rate (WER) of 5.06\%. These results highlight the effectiveness of preprocessing in improving OCR performance for historical datasets, despite challenges like variability in print quality.\\
The second phase involved transforming OCR-generated text into structured JSON files using AI. Using GPT-3.5 Turbo and a Pydantic-based schema as decoding constraints, we achieved an average accuracy of 65\% for JSON files created from correct text files (WER=0\%) and 63\% from OCR-generated text files (WER=5\%). These findings indicate that a Word Error Rate of 5\% does not significantly impact the accuracy of structured data extraction as expected.\\
In the third phase, we developed a record linkage algorithm to integrate extracted data into a centralized database. The algorithm demonstrated high accuracy (94\%) when working with manually curated JSON files but faced challenges with OCR-derived files (81\%), underscoring the impact of OCR errors on downstream processes. Enhancements such as improved matching criteria or name variation handling could address these limitations.\\
This work contributes to the field of digital humanities by showing the potential of AI-driven methods for preserving and enriching historical records. The framework developed here addresses common challenges in digitization, such as inconsistencies in layout and terminology, while providing a scalable solution for database integration. By combining OCR and AI technologies, this research lays the groundwork for future advancements in the digitization of cultural heritage.

\printbibliography


\appendix
\section{Example Person: OCR-Generated Text}\label{appendixF}
\begin{lstlisting}[language=json,firstnumber=1, caption={Text file generated using a trained Tesseract OCR engine, detailing the education and career of Franciscus (Francois) GOMARUS.}]
9

GOMARUS (GOMAIR), Franciscus (Francois)

Geb. Brugge 30-01-1563 (14)
Gest. Groningen 11-01-1641 (14)

Opleiding:
Stud. Litt., Phil., en Theol. Straatsburg 1577 (a,33)
Carriere:
Hoogleraar Theol. Leiden 25-01-1594 (14)
Nevenfuncties: (6)
Revisor Bijbelvertaling Syn. Den Haag 1598
Praeses Classis Vlissingen 1612 (a)
Afgev. Univ. Groningen bij
Synode Dordrecht 1618
Echtgenotes:

1. Anna Emerentia Musenhole (Muysenhol) (6,a)

Getr. Frankfurt a/d Main 1588 (a)

Gest. 1592 (54)

Vader: Gilles Muysenhol uit Antwerpen (a)

Kinderen: (a)

1. Franciscus

Geb. 1594

Getr. 1620 Agneta Wermeri
Getr. 1622 Maria Nissingh
Ouders:

Franciscus (Fransoys) Gomarus (4a,6)
Gegoede koopman, eigenaar herberg

Johanna Moermans (6)
Bijzonderheden:

Salaris: bij aanvang 7 800 (6)

a) G.P. van Ttterzon, Franciscus Gomarus 's-Gravenhage, 1930.
\end{lstlisting}
\section{Example Person: AI-Generated JSON}\label{appendixB}

\begin{lstlisting}[language=Python, caption={GPT prompt for extracting structured data from OCR text using GPT-3.5 Turbo}]

def chat_completion(person_info):

    return client.chat.completions.create(
        model="gpt-3.5-turbo",
        messages=[
            {
                "role": "system",
                "content": '''You are an advanced data extraction system.
                              - You can identify each person by surname
                              - The surname is always in uppercase letters, followed by the middle and/or first name
                              - If you can't determine the field value, refer to the examples'''
            },
            {
                "role": "user",
                "content": f'Please extract the data for the following person: {person_info}'
            }
        ],
        response_model=Person,
        max_retries=3,
        tool_choice="auto"
    )

\end{lstlisting}
\section{Example Person: AI-Generated JSON}\label{appendixG}
Listing 10: This JSON file, generated using GPT-3.5 Turbo based on OCR-generated text as input, provides information about Franciscus (Francois) GOMARUS.
\begin{multicols}{2}
\begin{lstlisting}[language=json,firstnumber=1], caption={}]
{
  "FirstName": "Franciscus (Francois)",
  "LastName": "GOMARUS",
  "Affix": "(GOMAIR)",
  "Gender": "Man",
  "second_names": [
    "Gomair"
  ],
  "alternative_last_names": [],
  "education": [
    {
      "subject": "Stud. Litt., Phil., en Theol.",
      "location": "Straatsburg",
      "date": "1577",
      "source": "33"
    }
  ],
  "careers": [
    {
      "job": "Hoogleraar Theol.",
      "location": "Leiden",
      "date": "1594-01-25",
      "source": "14",
      "is_side_job": 0
    }
  ],
  "particularities": [
    {
      "particularity": "Salaris: bij aanvang 7 800",
      "location": "s 'Gravenhage",
      "date": null,
      "source": "6"
    }
  ],
  "spouses": [
    {
      "FirstName": "Anna Emerentia",
      "LastName": "Musenhole",
      "Affix": null,
      "Gender": null,
      "source": "6,a",
      "second_names": [],
      "alternative_last_names": [],
      "BirthCountry": "Duitsland",
      "BirthCity": "Frankfurt",
      "BirthDate": null,
      "DeathDate": "1592",
      "DeathCity": null
    }
  ],
  "parents": [
    {
      "FirstName": "Franciscus",
      "LastName": "GOMARUS",
      "Affix": null,
      "Gender": null,
      "source": null,
      "second_names": [],
      "alternative_last_names": [],
      "BirthCountry": null,
      "BirthCity": null,
      "BirthDate": null,
      "DeathDate": null,
      "DeathCity": null
    }
  ],
  "grand_parents": [],
  "in_laws": [],
  "children": [
    {
      "FirstName": "Franciscus",
      "LastName": "GOMARUS",
      "Affix": null,
      "Gender": null,
      "source": null,
      "second_names": [],
      "alternative_last_names": [],
      "BirthCountry": null,
      "BirthCity": null,
      "BirthDate": "1594",
      "DeathDate": null,
      "DeathCity": null
    }
  ],
  "far_family": [],
  "type_of_person": 1,
  "faculty": "Theologie",
  "BirthCountry": null,
  "BirthCity": "Brugge",
  "BirthDate": "1563-01-30",
  "DeathDate": "1641-01-11",
  "DeathCity": "Groningen"
}

\end{lstlisting}
\end{multicols}

\end{document}